\documentclass[11pt]{article}
\usepackage{nodalida2021}
\usepackage{times}
\usepackage{url}
\usepackage{latexsym}

\usepackage{graphicx}
\usepackage{ tipa }

\usepackage[T2A,T1]{fontenc}
\usepackage[utf8]{inputenc}
\usepackage[russian,english]{babel}

\aclfinalcopy 

\title{Neural Morphology Dataset and Models for Multiple Languages, from the Large to the Endangered}

\author{Mika Hämäläinen, Niko Partanen, Jack Rueter and Khalid Alnajjar \\
  Faculty of Arts \\
  University of Helsinki \\
  {\tt firstname.lastname@helsinki.fi} \\ }

\date{}

\begin{document}
\maketitle
\begin{abstract}
  We train neural models for morphological analysis, generation and lemmatization for morphologically rich languages. We present a method for automatically extracting substantially large amount of training data from FSTs for 22 languages, out of which 17 are endangered. The neural models follow the same tagset as the FSTs in order to make it possible to use them as fallback systems together with the FSTs. The source code\footnote{https://github.com/mikahama/uralicNLP/wiki/Neural-morphology}, models\footnote{http://doi.org/10.5281/zenodo.3926769} and datasets\footnote{http://doi.org/10.5281/zenodo.3928628} have been released on Zenodo. 
\end{abstract}

\section{Introduction}

Morphology is a powerful tool for languages to form new words out of existing ones through inflection, derivation and compounding. It is also a compact way of packing a whole lot of information into a single word such as in the case of the Finnish word \textit{hatussanikinko} (in my hat as well?). 
This complexity, however, poses challenges for NLP systems, and in the work concerning endangered languages, morphology is one of the first NLP problems people address.

The GiellaLT infrastructure \cite{Moshagen2014} has HFST-based \cite{linden2013hfst} finite-state transducers (FSTs) for several morphologically rich (and mostly Uralic) languages. These FSTs are capable of lemmatization, morphological analysis and morphological generation of different words.

These transducers are at the core of this infrastructure, and they are in use in many higher level NLP tasks, such as rule-based \cite{trosterud2004porting} and neural disambiguation \cite{ens2019morphosyntactic}, dependency parsing \cite{antonsen2010reusing} and machine translation \cite{pirinen2017north}. 
The transducers are also in constant use in several real world applications such as online dictionaries \cite{rueter2019xml}, spell checkers \cite{skolt-spell-check}, online creative writing tools \cite{hamalainen2018poem}, automated news generation \cite{alnajjar2019no}, language learning tools \cite{antonsen2018using} and documentation of endangered languages \cite{gerstenberger2017instant,wilbur2018a}. As an additional important application we can mention the wide use of FSTs in the creation of Universal Dependencies treebanks for low-resource languages, at least with Erzya \cite{rueter2018towards}, Northern Saami \cite{tyers2017annotation} Karelian \cite{pirinen2019building} and Komi-Zyrian \cite{partanen2018first}. 

Especially in the context of endangered languages, accuracy is a virtue. Rule-based methods not only serve as NLP tools but also as a way of documenting languages in a machine-readable fashion. Members of language communities do not benefit, for example, from a neural spell checker that works to a degree in a closed test set, but fails miserably in real world usage. On the contrary, a rule based description of morphology can only go so far. New words appear and disappear all the time in a language, and keeping up with that pace is a never ending job. This is where neural models come in as they can learn to generalize rules for out-of-vocabulary words as well.
Pirinen \shortcite{pirinen2019a} also showed recently that at least with Finnish the neural models do outperform the rule-based models. This said, Finnish is already a larger language, so the experience doesn't necessarily translate into low-resource scenario (see \citealt{mika-endangered}).

The purpose of this paper is to propose neural models for the three different tasks the GiellaLT FSTs can handle: morphological analysis (i.e. given a form such as \textit{kissan}, produce the morphological reading \textit{+N+Sg+Gen}), morphological generation (i.e. given a lemma and a morphology, generate the desired form such as \textit{kissa+N+Sg+Gen} to \textit{kissan}) and lemmatization (i.e. given a form, produce the lemma such as \textit{kissan} to \textit{kissa} 
`a cat'). The goal is not to replace the FSTs, but to produce neural fallback models that can be used for words an FST does not cover. 
This way, the mistakes of the neural models can easily be fixed by fixing the FST, while the overall coverage of the system increases by the fact that a neural model can cover for an FST.

The main goal of this paper is not to propose a state of the art solution in neural morphology. 
The goal is to first build the resources needed to train such neural models so that they will follow the same morphological tags as the GiellaLT FSTs, and secondly train models that can be used together with the FSTs. 
All of the trained models will be made publicly available in a Python library that supports the use of the neural models and the FSTs simultaneously. 
The dataset built in this paper and the exact train, validation and test splits used in this paper have been made publicly available for others to use on the permanent archiving platform Zenodo. 

\section{Constructing the Dataset}

We are well aware of the existence of the popular UniMorph dataset \cite{mccarthy2020unimorph}. 
However, it does not suit our needs of two reasons. One reason is the incompatible morphological tagset. 
Our goal is to build models that can directly be used side-by-side with the existing FSTs, which means that the data has to follow the same formalism. 
Conversion is not a possibility, as the main reason we are not interested in using the UniMorph data is its limited scope; not only does it not cover all the languages we are dealing with in this paper, but it does not cover any cases of complex morphology. 
For example, the Finnish dataset does not cover possessive suffixes, question markers, comparative, superlative etc. 
Such a data would not be on par with the output produced by the FSTs.

We produce the data for the following languages: German (deu), Kven (fkv), Komi-Zyrian (kpv), Mokhsa (mdf), Mansi (mns), Erzya (myv), Norwegian Bokmål (nob), Russian (rus), South Sami (sma), Lule Sami (smj), Skolt Sami (sms), Võro (vro), Finnish (fin), Komi-Permyak (koi), Latvian (lav), Eastern Mari (mhr), Western Mari (mrj), Namonuito (nmt), Olonets-Karelian (olo), Pite Sami (sje), Northern Sami (sme), Inari Sami (smn) and Udmurt (udm). 
A vast majority of these languages are greatly endangered \cite{moseley_2010}.

\begin{table*}[]
\centering
\scriptsize
\tabcolsep=0.07cm
\begin{tabular}{|l|l|l|l|l|l|l|l|l|l|l|l|l|l|l|l|l|l|l|l|l|l|l|}
\hline
    & deu  & fin   & fkv  & koi & kpv   & lav  & mdf   & mhr   & mns  & mrj  & myv   & nob   & olo  & rus   & sje  & sma  & sme   & smj  & smn   & sms   & udm   & vro  \\ \hline
N   & 8741 & 51916 & 5936 & 558 & 20042 & 9738 & 17196 & 14079 & 2263 & 2529 & 10234 & 32009 & 5942 & 24691 & 2685 & 5946 & 37943 & 4331 & 13826 & 21158 & 10722 & 4703 \\ \hline
Adv & 588  & 6036  & 652  & 89  & 2942  & 953  & 1771  & 2346  & -    & 444  & 743   & 1743  & 14   & 2546  & -    & 543  & 1314  & 343  & 1146  & 1729  & 985   & 122  \\ \hline
V   & 4021 & 27875 & 1445 & 532 & 12504 & 2601 & 11983 & 9954  & 4924 & 2456 & 3781  & 7432  & 2782 & 14348 & 1751 & 5208 & 7724  & 3130 & 5436  & 5033  & 3669  & 4129 \\ \hline
A   & 2768 & 13056 & 917  & 128 & 5218  & 1652 & 4407  & 5116  & -    & 1031 & 2926  & 3236  & 2134 & 11054 & 185  & 645  & 2927  & 468  & 2295  & 3898  & 1550  & 1019 \\ \hline
\end{tabular}
\caption{The sizes of the GiellaLT dictionaries per part-of-speech}
\label{tab:lexicon-size}
\end{table*}

We use the FSTs  and dictionaries from the GiellaLT with the UralicNLP \cite{uralicnlp_2019} library to build the datasets for training the models. 
We do this in a clever way by taking all open class part-of-speech words from the dictionaries for each language and use the FSTs to produce all morphological readings for them. 
The number of words in the GiellaLT dictionaries is shown in Table \ref{tab:lexicon-size}. 
The FSTs do not let us do this by default, so we build a regular expression transducer that finds all possibilities for an input word and its part-of-speech.
In order to build the regular expression, we query all alphabets in the transducer that contain one of the following strings for exclusion: \textit{\#}, \textit{Der}, \textit{Cmp} or \textit{Err}. This will remove compounds, erroneously spelled forms and derivations. Derivations need to be excluded because otherwise the transducers would produce derivations of derivations of derivations and so on. 
Once the regular expression transducer is composed with the FST analyzer, we can use HFST to extract the transducer paths to get a list of all the possible morphological forms of the input word. 
From these, we filter out \textit{Clt} and \textit{Foc} tags because these multiply the number of possible morphological forms, especially since multiple different clitics can be appended after each other, and some times even in multiple different orders. 
We also remove tags indicating non-standard forms, \textit{Use} and \textit{Dial}, and \textit{Sem} tags that are used in language learning tools as well as contextual disambiguation to categorize semantically similar words. 
Table \ref{tab:unique-forms} shows how many unique inflectional forms each part-of-speech category has per language.

\begin{table*}[]
\centering
\footnotesize
\tabcolsep=0.07cm
\begin{tabular}{|l|c|c|c|c|c|c|c|c|c|c|c|c|c|c|c|c|c|c|c|c|c|c|}
\hline
    & deu & fin  & fkv & koi & kpv & lav  & mdf & mhr & mns & mrj & myv & nob & olo & rus & sje & sma & sme  & smj & smn & sms & udm & vro \\ \hline
N   & 24  & 850  & 50  & 788 & 183 & 24   & 83  & 208 & 151 & 162 & 19  & 17  & 98  & 75  & 16  & 50  & 727  & 297 & 496 & 339 & 744 & 26  \\ \hline
Adv & 1   & 16   & 1   & 2   & 4   & -    & 4   & 3   & -   & 2   & 2   & 2   & -   & 1   & -   & 3   & 8    & 3   & 2   & 3   & 1   & 6   \\ \hline
V   & 254 & 6667 & 139 & 198 & 249 & 1245 & 894 & 59  & -   & 40  & 10  & 21  & 726 & 693 & 38  & 58  & 302  & 144 & 382 & 177 & 156 & 119 \\ \hline
A   & 150 & 1244 & 77  & 4   & 244 & 44   & 127 & 4   & -   & 2   & 5   & 15  & 217 & 39  & 52  & 75  & 1347 & 187 & 100 & 627 & 54  & 100 \\ \hline
\end{tabular}
\caption{Number of unique inflectional forms per part-of-speech category}
\label{tab:unique-forms}
\end{table*}

We use the method described above to produce the data with all the open class part-of-speech words in the GiellaLT dictionaries for each language. 
For languages with bigger dictionaries, the maximum number of lemmas used per part of speech is set to 2100, in which case the lemmas are also picked at random. 
We use the typical split ratio and split 70\% of the data for training, 15\% for validation and 15\% for testing. 
The split is done on the lemma level and for each part-of-speech separately. 
This means that the test and validation sets will consist exclusively of out of vocabulary words that have not appeared in the training in any inflectional from. 
This also means that the ratios are the same for each part-of-speech, 70\% of the adjectives are used in the training, 70\% of the verbs and so on. 
The actual sizes can be seen in Table \ref{tab:data-splits}.

\begin{table*}[]
\centering
\scriptsize
\tabcolsep=0.07cm
\begin{tabular}{|l|l|l|l|l|l|l|l|l|l|l|l|l|l|l|l|l|l|l|l|l|l|l|}
\hline
           & deu    & fin      & fkv    & koi    & kpv    & lav    & mdf     & mhr    & mns    & mrj    & myv   & nob   & olo     & rus    & sje   & sma    & sme    & smj    & smn     & sms     & udm    & vro    \\ \hline
train      & 394k & 14486k & 286k & 483k & 873k & 320k & 1267k & 666k & 283k & 232k & 45k & 37k & 1054k & 243k & 80k & 108k & 799k & 648k & 1167k & 2831k & 943k & 257k \\ \hline
val & 87k  & 3061k  & 62k  & 105k & 186k & 68k  & 276k  & 142k & 60k  & 50k  & 9k  & 8k  & 229k  & 51k  & 17k & 22k  & 177k & 145k & 249k  & 628k  & 202k & 54k  \\ \hline
test       & 84k  & 3109k  & 60k  & 105k & 186k & 68k  & 274k  & 142k & 60k  & 50k  & 9k  & 8k  & 221k  & 53k  & 16k & 23k  & 179k & 143k & 253k  & 624k  & 203k & 55k  \\ \hline
\end{tabular}
\caption{Sizes of the datasets for each language. The splits do not share vocabulary.}
\label{tab:data-splits}
\end{table*}

The reason why we do the testing purely on out-of-vocabulary words is simply to test the accuracy of the models in the scenario that is more close to the one they are trained for, namely, in cases where the FSTs fail in their coverage.

\section{Experiments and Results}

In this section, we cover the neural architecture for the three separate morphological tasks: lemmatization, analysis and generation. We also show the results of the models in these tasks for each language, and present an error analysis on the Finnish and Komi-Zyrian by taking a closer look at the results.

\subsection{The Neural Model}

Over recent years, there has been a growing body of work on different neural approaches for low resourced languages in morphological analysis \cite{moeller-etal-2019-improving,schwartz-etal-2019-bootstrapping}, lemmatization \cite{kondratyuk-2019-cross,silfverberg2019data} and generation \cite{oseki-etal-2019-inverting,yu-etal-2020-ensemble}. 
Most notably the use of bi-directional LSTM architecture seems to be supported by most of the recent related work for analysis, generation and lemmatization. 

It is important to note that we approach the lemmatization and analysis from the same point of view as the FSTs. 
This means that it is a strictly morphological process, and the question of disambiguation is left for another part of the GiellaLT NLP pipeline, namely constraint grammar rules \cite{bick2015cg}. 
There is a plethora of work dealing with in-context lemmatization \cite{manjavacas-etal-2019-improving,malaviya-etal-2019-simple}, morphological analysis \cite{lim2018multilingual,zalmout-habash-2020-joint} and part-of-speech tagging \cite{perl-etal-2020-low,hoya-quecedo-etal-2020-neural}, but that is not what we are aiming for. 
We are aiming for neural models that can be used to complement the already existing systems relying on the GiellaLT infrastructure.

For all three tasks, we train a character based bi-directional LSTM model \cite{hochreiter1997long} by using OpenNMT-py \cite{opennmt} with the default settings except for the encoder where we use a BRNN (bi-directional recurrent neural network) \cite{schuster1997bidirectional} instead of the default RNN (recurrent neural network) as BRNN has been shown to provide a performance gain in a variety of tasks. 
We use the default of two layers for both the encoder and the decoder and the default attention model, which is the general global attention presented by Luong et al. \cite{luong2015effective}. 

\begin{table*}[]
\centering
\begin{tabular}{|l|c|c|}
\hline
 & \multicolumn{1}{l|}{input} & \multicolumn{1}{l|}{output} \\ \hline
lemmatization & k a u n i i m p a n s a k o & k a u n i s \\ \hline
analysis & k a u n i i m p a n s a k o & A Comp Sg Gen PxSg3 Qst \\ \hline
generation & k a u n i s A Comp Sg Gen PxSg3 Qst & k a u n i i m p a n s a k o \\ \hline
\end{tabular}
\caption{Example of the training data for each task}
\label{tab:example-input}
\end{table*}

Table \ref{tab:example-input} shows an example of the input and output of the training data in each of the three different tasks. 
Words are split into characters on both the input and output side of the data. 
Different morphological tags are treated as separate tokens, this means that FST morphologies consisting of multiple tags such as \textit{N+Msc+Sg+Dat} are simply split by the plus sign. 
We train a separate model for each task, meaning that we train three different models for each language: one for lemmatization, analysis and generation. 
All models have shared the same random seed (3435), therefore training the models again with this seed should result in the exact same results we are reporting in this paper.

\subsection{Results}

We report the performance of the models in terms of accuracy, meaning how many results were fully right (entirely correct lemma, entirely correctly generated form and entirely correct morphological analysis). In addition, we report CER (character error rate) for the lemmatizers and generators, and a MER (morphological error rate) for the analyzers. 
These values indicate how close the model got to the correct result even if some of the results were a bit erroneous. 

\begin{table*}[]
\centering
\scriptsize
\tabcolsep=0.07cm
\begin{tabular}{|l|c|c|c|c|c|c|c|c|c|c|c|c|c|c|c|c|c|c|c|c|c|c|}
\hline
 & deu & fin & fkv & koi & kpv & lav & mdf & mhr & mns & mrj & myv & nob & olo & rus & sje & sma & sme & smj & smn & sms & udm & vro \\ \hline
gen acc & 0,65 & 0,64 & 0,68 & 0,67 & 0,78 & \textbf{0,95} & \textbf{0,85} & 0,58 & 0,78 & \textbf{0,90} & \textbf{0,93} & \textbf{0,94} & \textbf{0,83} & \textbf{0,97} & 0,77 & 0,69 & 0,73 & 0,67 & 0,57 & 0,40 & \textbf{0,87} & \textbf{0,82} \\
gen CER & 5,61 & 8,03 & 3,70 & 8,67 & 3,75 & 1,21 & 1,77 & 11,76 & 3,92 & 1,77 & 0,67 & 1,23 & 2,12 & 0,48 & 4,28 & 5,81 & 4,19 & 3,54 & 4,25 & 6,65 & 1,90 & 3,35 \\ \hline
lem acc & \textbf{0,88} & 0,68 & \textbf{0,80} & 0,70 & \textbf{0,87} & \textbf{0,85} & \textbf{0,93} & \textbf{0,88} & 0,79 & \textbf{0,88} & \textbf{0,90} & 0,76 & \textbf{0,87} & \textbf{0,82} & 0,72 & 0,71 & 0,06 & 0,70 & 0,67 & 0,79 & \textbf{0,92} & 0,79 \\
lem CER & 2,71 & 12,37 & 5,85 & 11,41 & 1,21 & 4,34 & 1,11 & 2,46 & 4,87 & 3,82 & 1,50 & 5,71 & 3,76 & 4,25 & 6,78 & 6,96 & 55,72 & 9,13 & 7,78 & 4,45 & 2,75 & 5,81 \\ \hline
ana acc & 0,11 & 0,57 & \textbf{0,86} & 0,78 & \textbf{0,88} & 0,39 & 0,61 & \textbf{0,94} & 0,77 & \textbf{0,92} & \textbf{0,98} & 0,49 & \textbf{0,86} & 0,36 & 0,73 & 0,60 & 0,56 & 0,53 & 0,42 & 0,42 & 0,76 & 0,74 \\
ana MER & 35,40 & 16,66 & 6,52 & 7,24 & 3,06 & 18,24 & 11,54 & 3,73 & 7,76 & 4,75 & 0,41 & 38,09 & 5,85 & 19,04 & 19,45 & 22,91 & 17,24 & 22,48 & 23,20 & 20,11 & 10,90 & 9,82 \\ \hline
\end{tabular}
\caption{Results of the models for different languages on out-of-vocabulary data}
\label{tab:results}
\end{table*}

The results can be seen in Table \ref{tab:results}, the models reaching to an accuracy to over 80 \% are highlighted in bold. 
The results indicate that lemmatization is the easiest task for the model to learn, and after that generation. 
Morphological analysis is the most difficult task as it receives the scores lower than the generation or lemmatization. 
Needless to say, some results are exceptionally good for specific languages such as for Erzya (myv) and Western Mari (mrj), while they are not good for others like Finnish (fin) and German (deu). 
This calls for more investigation of the results.

\begin{figure*}[h]
   \centering
   \includegraphics[page=1,width=1.0\textwidth]{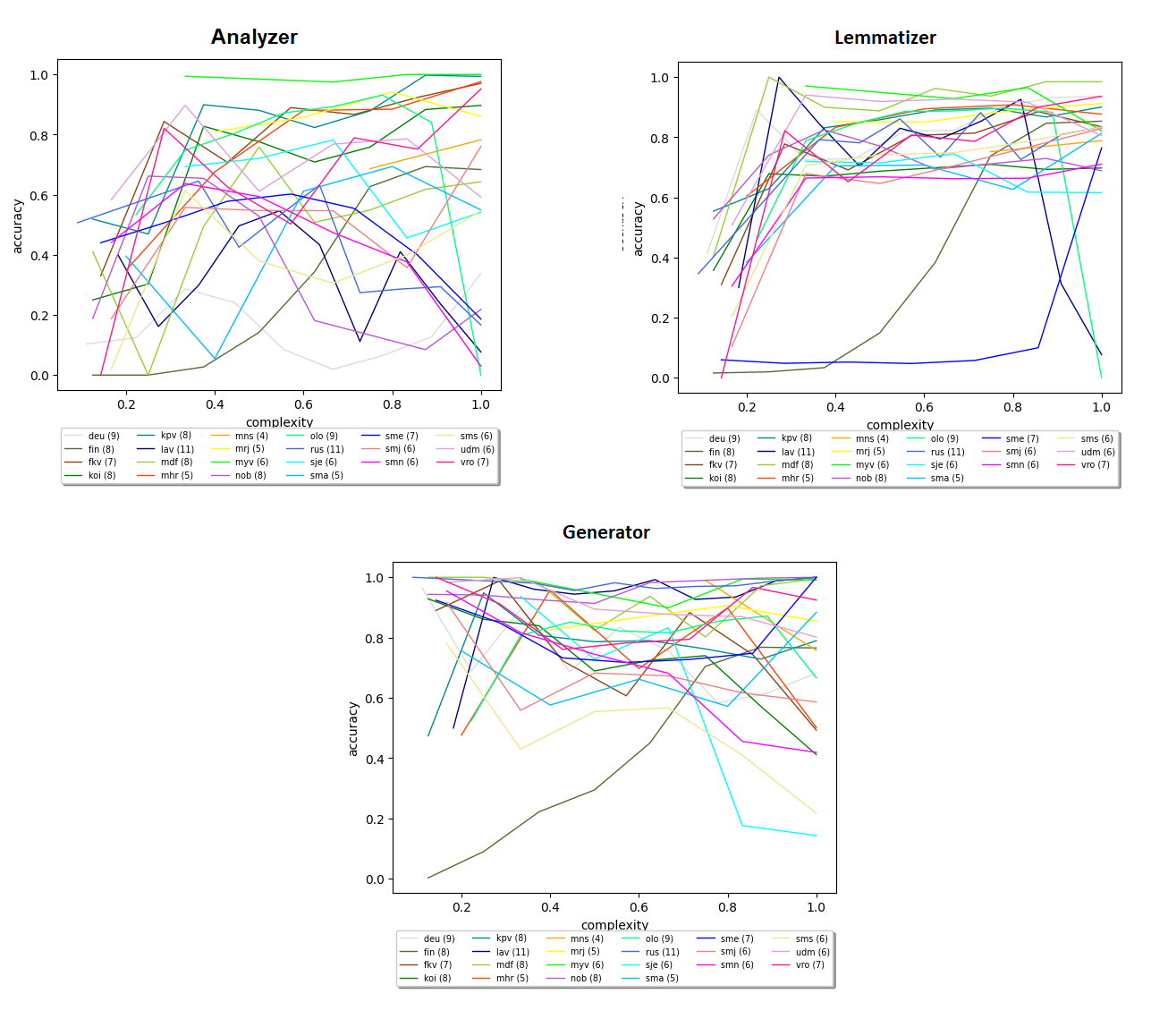}
 \caption{Accuracies based on morphological complexity}
 \label{fig:complexity}
\end{figure*}

Figure \ref{fig:complexity} shows the accuracy of each model based on the morphological complexity of the input. 
The complexity is measured by the number of morphological tags in the FST produced data. 
The complexity axis of the plots shows a relative complexity for each language, meaning that 1.0 has the maximum number of tags, 0.8 shows results for input having 80\% of the maximum number of tags and so on. 
The maximum complexity is shown in brackets after the language ISO-code. 
Analyzers seem to have a lower accuracy for most of the languages when the complexity is small. 
This is probably due to the fact that shorter word forms tend to have more ambiguity to begin with and might be analyzed as a word different from the one in the gold standard. 
For many languages, the accuracy increases towards the average complexity and drop again for the most complex forms. 
It is to be remembered that these accuracies are also affected by the peculiarities of the transducers themselves and their tagging conventions. 

Lemmatizers seem to follow the pattern of the analyzers but do so more clearly. 
Lemmatization of morpholgically simple forms is not as easy as more complex forms. 
However, as the complexity increases, the lemmatization accuracy does not drop for most of the languages. 
This has probably something to do with the fact that unlike morphological tags, the word forms follow clearer patterns as they do not have such a large amount of subjectivity in the tagging decisions the different linguists working on these transducers have introduced.

Generators are very even for most of the languages in the sense that they produce consistently around the same accuracy regardless of the morphological complexity. 
Although, some of the languages follow a more analyzer like pattern, generating wrong with small and large morphological complexity.

\begin{table*}[]
\small
\begin{tabular}{|l|l|l|}
\hline
    & Missing predictions                                    & Wrong predictions                                         \\ \hline
deu & Def, Pl, Acc, Dat, Neu, Gen, Msc, Fem, NoArt, Nom      & NoArt, Fem, Msc, Indef, Sg, Gen, Acc,   Nom, Def, Neu     \\ \hline
fin & PxSg3, A,   PxPl3, N, Sg, Pl, Nom, Gen, Par, Pss       & V,   Act, PxPl3, PrfPrc, PxSg3, Ind, PrsPrc, Pss, Prs, Sg \\ \hline
fkv & A, Act, N, Sg, V, Pl, Ind, Inf3, Nom, Pl3              & N, Act, A, V, Sg, Pl, Ind, Pass, Prs,   Inf3              \\ \hline
koi & IV, TV, AprIne,   AprIll, Ill, Prs, V, So/CP, Apr, Ind & Apr,   So/CP, TV, Ine, AprIne, Fut, Sg, N, IV, Nom        \\ \hline
kpv & Ine, TV, IV, Fut, Prs, Ill, V, Pl3, Sg1, PxSg1         & Ill, IV, TV, Prs, Ine, Fut, Sg, Sg3, N,   Pl1             \\ \hline
lav & IV, Fem, Acc,   Pl, Sg, Nom, TV, Voc, Def, Gen         & TV,   Gen, Msc, Sg, Pl, Indef, Loc, IV, Fem, Acc          \\ \hline
mdf & IV, TV, V, Ind, Prt2, OcPl3, N, A, PxSg2, NomAct       & TV, IV, N, Conj, OcSg3, Def, A, V,   OcPl1, ScSg1         \\ \hline
mhr & N, Sg, V,   So/CP, Nom, Ill, Ind, So/PC, Gen, Ger      & V,   Ind, N, Sg, Adv, Nom, Prs, Sg3, Imprt, A             \\ \hline
mns & PxPl2, Sg, Pl, PxDu2, Nom, PxSg2, Du, Lat, Abl, Loc    & PxDu2, Pl, Sg, PxSg2, Nom, PxPl3, Du,   Lat, Tra, PxPl2   \\ \hline
mrj & Sg, N, Lat,   Prs, V, Nom, Ind, Imprt, Ine, PxPl3      & Prt1,   Ind, V, Ill, N, Sg, Nom, Sg3, Ine, Prs            \\ \hline
myv & N, A, Tra, Ela, Abl, Ine, Interr, PxSg2                & A, N, Abl, Ine, Ela, Ill, Tra, NomAg, V,   IV             \\ \hline
nob & A, Pos, Indef,   Sg, Pl, PrfPrc, Def, Fem, V, MF       & Msc,   Ind, V, Sg, Indef, Prt, Neu, Def, N, Pl            \\ \hline
olo & Ins, N, A, ConNeg, V, Sg, Act, PrfPrc, Nom, Pl         & Gen, ConNeg, Act, V, N, Ind, A, Sg, Prs,   PrsPrc         \\ \hline
rus & TV, Acc, Neu,   Gen, Anim, Inan, Impf, Dat, Pass, IV   & IV,   Loc, Msc, AnIn, Nom, Perf, TV, Acc, Gen, Sg         \\ \hline
sje & Sg, V, Com, Prs, N, Sg2, Pl, Gen, Ind, Nom             & N, Sg, Pl, V, Ind, Nom, Prs, Prt, Gen,   Ine              \\ \hline
sma & IV, TV, N, Ind,   V, Ess, Sg, Pl, A, Prs               & Sg,   IV, TV, N, Ind, Com, Prs, Pl, Nom, Ill              \\ \hline
sme & Acc, Gen, TV, Sg, Pl, Loc, IV, N, Com, Nom             & Gen, Sg, Acc, IV, TV, Nom, Pl, Com, Loc,   A              \\ \hline
smj & Com, Sg, N, Pl,   PxDu2, PxDu1, Gen, Acc, IV, NomAg    & Gen,   TV, V, PxPl2, PxPl1, Pl, Sg, Nom, IV, Ine          \\ \hline
smn & Acc, Sg, Gen, PxPl2, PxSg3, N, PxPl3, IV, PxPl1, TV    & PxDu3, Gen, PxDu2, Nom, A, Ill, Sg,   PxSg1, PxSg2, Pl    \\ \hline
sms & N, Sg, V, Acc,   Pl, Gen, Ill, Com, Ind, Nom           & A,   Pl, Nom, Sg, Gen, Loc, Acc, Com, Ill, Par            \\ \hline
udm & Ill, N, Opt, Sg, ConNeg, Ind, PxSg3, Imprt, Sg2, Ine   & ConNeg, Ind, Ine, Sg3, Fut, Nom, N, A,   Det, Sg          \\ \hline
vro & A, Pss, Act, V,   Pl, Sg, Ind, N, Gen, ConNegII        & Sg,   Nom, Act, N, A, Pl, Sg1, Prs, Ind, Par              \\ \hline
\end{tabular}
\caption{The top 10 most difficult tags for the analyzers}
\label{tab:analysis-errors}
\end{table*}

Table \ref{tab:analysis-errors} shows the most difficult tags for the analyzers. 
The missing predictions column shows the most frequent tags the analyzer did not predict even though they were in the gold data, and the wrong predictions column shows the most frequent ones the analyzer predicted but were not in the gold data.
We can see that many of the most challenging tags are shared by different languages. 
In various Uralic languages, for example, connegatives and imperatives, or connegatives and infinitives, are homonymous, and cannot be predicted correctly just from the surface form alone.
Similarly cases such as illative and inessive are in many complex forms homonymous in Permic languages, which surfaces in missing predictions of all these languages.
In the languages where transitivity is a feature coded into FST, there are regular problems in predicting these categories correctly. 
Similarly, in many Indo-European languages gender is primarily a lexical category, and in many instances the model cannot predict it correctly in cases where only the surface form that doesn't show the gender is presented.
In the Section~\ref{sec:errors} we go through more in detail this kind of instances, for example, in relation to purely lexically determined Komi-Zyrian stem consonants.


Table \ref{tab:morph-complex} shows the morphological constructions that were the most difficult ones for the models to lemmatize and generate correctly in their respective columns. For instance, the Erzya (myv) generation indicates the translative with subsequent possessive-suffix marking is the most problematic. If it had been lemmatization, the explanation would point to the extreme infrequency of these translative forms and the fact that there is an ambiguity with genitive and nominative forms of derivations in \textit{ks}. Lemmatization for Erzya, however, appears to have no issues with ambiguity at all.
The same difficulties are not shared by other languages, but seem to all be language specific. Eastern and Meadow Mari (mhr), for example, appear to have difficulties with generation and lemmatization of nearly the same tag set, namely, the illative plural with a third person plural possessive suffix (ordered: possessive, plural and finally case marker). 
Looking at the sibling language Western Mari (mrj), we will note that there is a different tagging strategy in use, but here as well there seems to be an intersection where the same forms present problems for both generation and lemmatization.

This could be seen as a type of sanity test whereby simple flaws in the transducers might be detected. The Latvian (lav) transducer is a blatant example of inconsistencies in transducer development. 
The problem, which has  now been addressed and corrected, was in the multiple exponence of part-of-speech tags, i.e. there are double \textit{+V} and \textit{+N} tags due to the introduction of automated part-of-speech tagging in XML dictionary to FST formalism transformation without removing the part-of-speech tagging in subsequent continuation lexica of the rule-based transducer. Development of the Mari pair might be greatly enhanced through the introduction of a segment-ordering tag in Western or Hill Mari (mrj), which would bring it closer to the strategy followed in the Eastern and Meadow Mari (mhr) use of \textit{+So/PNC}. 
These questions with tag and suffix ordering appear also as important factor in Komi-Zyrian morphological generation, as discussed in Section~\ref{sec:errors}.

\begin{table*}[t]
\centering
\tiny
\tabcolsep=0.04cm
\begin{tabular}{|l|l|l|}
\hline
    & generator                                                                                                                                                                                                   & lemmatizer                                                                                                                                                                             \\ \hline
deu & \begin{tabular}[c]{@{}l@{}}V+PrfPrc+Pos+Pl+Nom+Indef, V+PrfPrc+Pos+Fem+Sg+Nom+Indef,  \\ V+PrfPrc+Pos+Pred, V+PrfPrc+Pos+Fem+Sg+Acc+Indef, \\ V+PrfPrc+Pos+Neu+Sg+Acc+Def\end{tabular}                      & \begin{tabular}[c]{@{}l@{}}N+Msc+Pl+Dat, N+Msc+Pl+Gen,   N+Msc+Pl+Nom, \\ N+Msc+Pl+Acc, N+Msc+Sg+Dat\end{tabular}                                                                      \\ \hline
fin & \begin{tabular}[c]{@{}l@{}}A+Sg+Ess+PxSg3,   A+Sg+Ess+PxPl3, \\ A+Sg+Ess+PxPl3+Qst, A+Sg+Ess+PxSg3+Qst, \\ N+Pl+Par+PxPl3+Qst\end{tabular}                                                                  & \begin{tabular}[c]{@{}l@{}}A+Sg+Ess+PxSg3,   A+Sg+Ess+PxPl3, A+Sg+Ess+PxPl3+Qst, \\ A+Sg+Ess+PxSg3+Qst, N+Pl+Par+PxSg3\end{tabular}                                                    \\ \hline
fkv & \begin{tabular}[c]{@{}l@{}}V+Act+Inf3+A+Pl+Superl+Par, A+Pl+Superl+Par, N+Pl+All,   \\ V+Act+Inf3+A+Pl+Par, V+Act+Inf3+A+Pl+Gen\end{tabular}                                                                & N+Pl+All, N+Pl+Par, N+Pl+Gen, N+Sg+Par,   N+Pl+Abe                                                                                                                                     \\ \hline
koi & \begin{tabular}[c]{@{}l@{}}V+Ind+Prt2+Pl3+Comp,   V+Ind+Prt2+Pl3, V+IV+Ind+Prt2+Pl3+Comp, \\ V+IV+Ind+Prt2+Pl3, V+TV+Ind+Prt2+Pl3\end{tabular}                                                              & \begin{tabular}[c]{@{}l@{}}N+Sg+Ela+Comp+Cop+Pl,   N+Sg+Ine+PxPl1+Comp+Cop+Pl, \\ N+Sg+Ine+PxPl2+Comp+Cop+Pl,   \\ N+Sg+Ine+PxPl3+Comp+Cop+Pl, N+Sg+Ela+PxSg1+Comp+Cop+Pl\end{tabular} \\ \hline
kpv & \begin{tabular}[c]{@{}l@{}}N+Sg+Com+PxSg2, N+Sg+Com+PxSg3, N+Sg+Egr+PxSg1+Comp,   \\ N+Sg+Egr+PxSg1, N+Sg+Egr+PxSg1+Comp+Cop+Pl\end{tabular}                                                                & \begin{tabular}[c]{@{}l@{}}N+Sg+Acc, Adv, N+Sg+Prl+PxPl1,   N+Sg+Com+PxSg2, \\ N+Sg+Com+PxSg3\end{tabular}                                                                             \\ \hline
lav & \begin{tabular}[c]{@{}l@{}}V+V+TV+PrsPrc+Act+Msc+Sg+Voc+Def,   \\ V+V+TV+PrsPrc+Pss+Msc+Sg+Voc+Def, \\ V+V+TV+PrfPrc+Pss+Msc+Sg+Voc+Def,   \\ N+N+Msc+Sg+Voc, V+V+IV+PrsPrc+Act+Msc+Sg+Voc+Def\end{tabular} & \begin{tabular}[c]{@{}l@{}}N+N+Msc+Sg+Voc,   N+N+Fem+Sg+Voc, N+N+Fem+Pl+Gen, \\ N+N+Fem+Sg+Acc, N+N+Fem+Sg+Loc\end{tabular}                                                            \\ \hline
mdf & \begin{tabular}[c]{@{}l@{}}V+IV+NomAg+Pl+Gen+PxSg3, V+IV+NomAg+Pl+Nom+PxSg3,   \\ V+IV+NomAg+SP+Cau+Indef, V+IV+NomAg+Sg+Dat+PxSg1, \\ V+IV+NomAg+Pl+Dat+PxSg3\end{tabular}                                 & \begin{tabular}[c]{@{}l@{}}N+Sg+Nom+Indef, N+SP+Gen+Indef,   N+Sg+Dat+PxSg2, \\ N+Sg+Dat+PxSg1, N+SP+Tra+Indef\end{tabular}                                                            \\ \hline
mhr & \begin{tabular}[c]{@{}l@{}}N+Pl+Ill+PxSg3+So/PNC,   N+Pl+Ill+PxPl1+So/PNC, \\ N+Pl+Ill+PxPl2+So/PNC, N+Pl+Ill+PxPl3+So/PNC,   \\ N+Pl+Ill+PxSg1+So/PNC\end{tabular}                                         & \begin{tabular}[c]{@{}l@{}}Adv,   N+Pl+Ill+PxSg3+So/PNC, N+Pl+Ill+PxSg2+So/PNC,\\  N+Sg+Ill+PxSg3+So/CP,   N+Pl+Ill+PxSg1+So/PNC\end{tabular}                                          \\ \hline
mns & \begin{tabular}[c]{@{}l@{}}N+Du+PxDu1+Abl, N+Du+PxDu1+Ins, \\ N+Du+PxDu1+Nom,   N+Du+PxDu1+Loc, \\ N+Du+PxDu1+Lat\end{tabular}                                                                              & \begin{tabular}[c]{@{}l@{}}N+Pl+PxSg2+Ins, N+Pl+PxSg2+Loc,   N+Pl+PxSg2+Abl, \\ N+Pl+PxSg2+Nom, N+Pl+PxSg2+Lat\end{tabular}                                                            \\ \hline
mrj & \begin{tabular}[c]{@{}l@{}}N+Sg+Ine+PxSg3,   N+Sg+Ill, N+PxSg2+Pl+Ill, \\ N+Sg+PxSg2+Ill, N+PxSg1+Pl+Lat\end{tabular}                                                                                       & \begin{tabular}[c]{@{}l@{}}N+Sg+Ill,   N+Sg+Gen, N+Sg+Acc, N+Sg+Nom, \\ N+Sg+Ine+PxSg3\end{tabular}                                                                                    \\ \hline
myv & \begin{tabular}[c]{@{}l@{}}N+SP+Tra+PxSg2, N+SP+Tra+PxPl1, N+SP+Tra+PxPl2,  \\  N+SP+Tra+PxPl3, N+SP+Tra+PxSg3\end{tabular}                                                                                 & \begin{tabular}[c]{@{}l@{}}V+IV+Act+PrsPrc,   V+IV+NomAg+SP+Ill+PxSg2, \\ V+IV+NomAg+SP+Ill+PxPl1, V+IV+NomAg+SP+Ill+PxSg3,   \\ V+IV+NomAg+SP+Ine+PxPl1\end{tabular}                  \\ \hline
nob & \begin{tabular}[c]{@{}l@{}}N+Neu+Pl+Def,   V+Ind+Prt, N+Neu+Pl+Indef, \\ V+PrfPrc, A+Superl+Def\end{tabular}                                                                                                & \begin{tabular}[c]{@{}l@{}}V+Imp,   A+Pos+Neu+Sg+Indef, A+Pos+Fem+Sg+Indef, \\ V+Ind+Prt, A+Pos+Msc+Sg+Indef\end{tabular}                                                              \\ \hline
olo & \begin{tabular}[c]{@{}l@{}}V+Act+PrsPrc+Pl+Abe, V+Act+PrsPrc+Pl+Abe+Qst, \\ N+Pl+Abe,   N+Pl+Abe+Qst, N+Sg+Abe+Qst\end{tabular}                                                                             & \begin{tabular}[c]{@{}l@{}}N+Sg+Nom, N+Sg+Nom+Qst, N+Sg+Abe,   \\ N+Pl+Abe+Qst, N+Sg+Abe+Qst\end{tabular}                                                                              \\ \hline
rus & \begin{tabular}[c]{@{}l@{}}V+Perf+IV+Imp+Pl2,   V+Perf+IV+Imp+Sg2, \\ V+Perf+IV+Fut+Sg3, V+Perf+IV+Fut+Sg2, V+Perf+IV+Fut+Sg1\end{tabular}                                                                  & \begin{tabular}[c]{@{}l@{}}Adv,   A+Neu+Sg+Pred, A+Msc+Sg+Pred, \\ V+Perf+IV+Imp+Sg2, A+Msc+AnIn+Sg+Loc\end{tabular}                                                                   \\ \hline
sje & N+Pl+Ela, N+Pl+Com, N+Sg+Ela, N+Sg+Com, V+Pot+Sg3                                                                                                                                                           & N+Pl+Com, N+Pl+Ela, N+Sg+Ela, N+Sg+Com,  V+Ind+Prs+Sg3                                                                                                                                 \\ \hline
sma & N+Pl+Gen,   N+Pl+Com, N+Sg+Com, N+Pl+Ill, N+Ess                                                                                                                                                             & N+Pl+Gen,   N+Sg+Gen, N+Ess, N+Pl+Ine, N+Pl+Nom                                                                                                                                        \\ \hline
sme & \begin{tabular}[c]{@{}l@{}}A+Comp+Sg+Nom+Qst, A+Comp+Sg+Nom, \\ A+Comp+Attr,   A+Comp+Attr+Qst, V+TV+VAbess+Qst\end{tabular}                                                                                & \begin{tabular}[c]{@{}l@{}}A+Comp+Sg+Nom+Qst, A+Comp+Sg+Nom,   \\ A+Comp+Attr, A+Comp+Attr+Qst, V+TV+VAbess\end{tabular}                                                               \\ \hline
smj & \begin{tabular}[c]{@{}l@{}}N+Sg+Com+PxSg1,   N+Sg+Com+PxSg2, \\ N+Sg+Abe, N+Pl+Abe, N+Pl+Gen+PxSg1\end{tabular}                                                                                             & \begin{tabular}[c]{@{}l@{}}N+Pl+Abe,   N+Sg+Abe, N+Sg+Com+PxSg1, \\ N+Sg+Com+PxSg2, N+Pl+Com+PxSg2\end{tabular}                                                                        \\ \hline
smn & \begin{tabular}[c]{@{}l@{}}N+Pl+Com+Qst, A+Pl+Com+Qst, A+Comp+Pl+Com+Qst,  \\  A+Superl+Pl+Com+Qst, V+PrsPrc+Qst\end{tabular}                                                                               & \begin{tabular}[c]{@{}l@{}}N+Pl+Com+Qst, N+Pl+Gen+Qst,   V+Ind+Prs+Sg3+Qst, \\ A+Pl+Com+Qst, V+Ind+Prs+ConNeg+Qst\end{tabular}                                                         \\ \hline
sms & \begin{tabular}[c]{@{}l@{}}A+Superl+Sg+Abe,   A+Superl+Sg+Abe+Qst/a, \\ A+Superl+Sg+Abe+Qst/ko, V+VAbess+Qst/a,  V+VAbess+Qst/ko\end{tabular}                                                               & \begin{tabular}[c]{@{}l@{}}V+Ind+Prt+Pl1,   V+VAbess+Qst/a, V+Ind+Prt+Pl1+Qst/ko, \\ V+VAbess+Qst/ko, V+VAbess\end{tabular}                                                            \\ \hline
udm & \begin{tabular}[c]{@{}l@{}}N+Sg+Ela+PxPl1, N+Sg+Ela+PxSg3, \\ N+Sg+Ela+PxSg2,   N+Sg+Ela+PxPl3+Qst, N+Sg+Ela+PxSg1\end{tabular}                                                                             & \begin{tabular}[c]{@{}l@{}}V+Ind+Prs+Pl1, V+Ind+Prs+Pl1+Qst,  \\ V+Ind+Fut+Pl1+Qst, V+Ind+Fut+Pl1, V+Imprt+Pl2\end{tabular}                                                            \\ \hline
vro & \begin{tabular}[c]{@{}l@{}}V+Act+Sup+Ine,   V+Act+Ind+Prt+Sg2, V+Pss+Ind+Prt+Sg2, V\\ +Pss+PrfPrc, V+Pss+PrfPrc+Sg+Nom\end{tabular}                                                                         & \begin{tabular}[c]{@{}l@{}}V+Act+Ind+Prt+Sg2,   N+Pl+Ill, V+Pss+PrfPrc, \\ V+Pss+PrfPrc+Sg+Nom, V+Pss+Ind+Prt+Sg2\end{tabular}                                                         \\ \hline
\end{tabular}
\caption{The top 5 morphological forms that were the most difficult to lemmatize and generate}
\label{tab:morph-complex}
\end{table*}

\subsection{Error Analysis}
\label{sec:errors}

In this section, we take a closer look at the result of the Finnish (fin) and Komi-Zyrian (kpv) models in order to better understand their shortcomings.

\subsubsection{Finnish}

For lemmatization Finnish offered one of the worst results, which makes it an interesting target for error analysis.
Some of the obvious errors are related to extremely common word formation patterns, which the model for some reason is not able to generalize.
One of these pattern belongs to adjectives and nouns formed with suffix -inen, for example \textit{pienimuotoisissani} `in my most minor (things)' the correct lemmatization would be \textit{pienimuotoinen}, but the model returns \textit{pienimuotoida}, which doesn't mean anything. 
Interestingly, it gives very consistently similar forms to different variants of the same word, so the model appears to believe this is the correct lemma.
We can analyze that out of all Finnish lemmatization errors -inen derivations are involved in 7.7\% of all mistakes. 
Thereby future work should investigate what can cause such a gap in the models prediction abilities, as impact in this can lead into rapid improvements.
One phenomena we observed is that Finnish FST also produces incorrect forms, such as \textit{pienimuotoisimmillean}, which probably should end into \textit{-een}. 
We can also observe that in many Finnish lemmas that the model does analyze correctly the forms are compounds.
This leaves open the possibility that the training data has contained either the second component independently or within a comparable compound, which would had given the model some example.
One lemmatization issue that can be distinguished is that the model doesn't lemmatize correctly proper names that are written with initial capital letter.
These include several words, for example \textit{Unkareinansako} `as their Hungaries?' should be lemmatized as \textit{Hungary}, but the model returns \textit{nkareintaa}. 
What this shows is that the model struggles with uppercase characters, although those would ideally be part of the correct lemmatization result.

The Finnish model has problems in generating forms for words ending in \textit{-lainen}, as it seems to inflect them as one would inflect the word \textit{laine} `wave', such as \textit{dominikaanilaineiltasi} `$\approx$ from your Dominican waves' instead of \textit{dominikaanilaisiltasi} `from your Dominican people'. Also, other adjectives ending in \textit{-inen} are problematic such as \textit{keväneensä} instead of \textit{keväisensä} 
`his spring-like'. In this case, the model has not learned the typical inflectional category of adjectives ending in \textit{-inen}.
This issue has an interesting parallel with the same problem being present in the lemmatization task, described above.
This shows that the problems the models encounter are to some degree parallel to one another in different tasks, and either relate to the complexity of the linguistic system, or somehow inadequately represented input.

Interestingly, the generation model has problems with the plural forms of the abessive and illative case, and often generates the singular form instead of the plural such as in \textit{sähkömittariksesi} `for your electricity meter' instead of \textit{sähkömittareiksesi} `for your electricity meters' or a completely erroneous form such as \textit{sähkömittaritsiisi} instead of \textit{sähkömittareihisi} 
`to your electricity meters'. 
In these erroneous cases, the model has tried to pluralize the word, for example \textit{sähkömittarit} is the correct plural form of electricity meters in nominative, but it is no longer correct when inflected in the illative case.

\subsubsection{Komi-Zyrian}

When we examine the lemmatization task, some particularities are obvious in Komi-Zyrian. 
For example, many of word forms with interspersed white spaces in them are not lemmatized correctly.
We also see that some complex entries borrowed from Russian are challenging to lemmatize, possibly due to their rarity, for example: \foreignlanguage{russian}{народно-освободительнӧйджыкъяснысланьджык} 
`more in the direction of their people who are more national-liberational'
would correctly result in
\foreignlanguage{russian}{народно-освободительнӧй}, but the model predicts 
\foreignlanguage{russian}{народнотильнӧй}.
In this case the hyphen within the compound probably contributes to the rarity of the form itself. 
Similarly, the model is also struggling when there are words that follow orthographic conventions more typical to Russian than Komi, for example 
\foreignlanguage{russian}{областьсаас}
would be correctly lemmatized as \foreignlanguage{russian}{областьса}, 
but the model predicts \foreignlanguage{russian}{областььса}.
If this reflects the underlying code, model training like this could be very useful for locating erroneously coded transducers. The double soft sign would seem to allude to double exponence in the code.
The model also has challenges with rarer orthographical conventions in Komi vocabulary. For example \foreignlanguage{russian}{пипуа-кыддзаинӧйланьсянь} 
`from the direction of my aspen and birch grove' 
should be \foreignlanguage{russian}{пипуа-кыддзаин} 
`aspen and birch grove', but we get \foreignlanguage{russian}{пипуа-кыдзаин}. These shortcomings, however, are relatively rare in the Zyrian data, and the model learns to lemmatize at high accuracy. 
Much more so than Finnish, which could be related to more concatenative morphology of Komi where the word boundaries can be easier to detect.

In the case of Komi-Zyrian we can observe that a large portion of wrongly recognized forms results from ambiguity that is inherent to the morphology of this language.
For example, it is not possible to distinguish some of the cases, such as the inessive and illative, in all forms where they occur.
As the model inevitably returns only one reading, it is clear that the evaluation accuracy cannot be perfect. This finding is consistent with analogous ambiguity for other forms in the Skolt Sami (sms) model. 
There appears to be a consistency in what is incorrectly predicted in Skolt Sami. 
When there is a four-way ambiguity as in the \textit{Sg Gen}, \textit{Sg Acc} \textit{Sg Nom} and \textit{Pl Nom}, the tag \textit{Sg Gen} is consistently predicted to be \textit{Pl Nom}, leaving the two readings \textit{Sg Acc} and \textit{Sg Nom} out of the dichotomy.
Komi models shows similar preferences into specific categories when there are multiple homonymous possibilities.

In the analysis above it was already briefly discussed that some categories are difficult to recognize correctly for Permic languages.
Another example like this is seen in the Komi-Zyrian and Komi-Permyak (koi) future tense marking. 
As these languages have morphologically marked future in the third person alone, every first and second person verb in the present tense also gets a future reading, as both analyses can be seen as correct.
One could also argue, however, that if some analysis is not possible to resolve at this level, some of the distinctions could be removed or merged at this level of analysis. 

What comes to morphological generation of Komi, the accuracy is rather high. 
Some of the errors can be connected to the fact that some suffixes can occur in varying orders.
For example, with input \foreignlanguage{russian}{кольквиж A Sg Egr PxPl1 Comp} 
one could assume the output \foreignlanguage{russian}{кольквижнымсяньджык} 
`more from the direction of our yellows', but in this case the model outputs \foreignlanguage{russian}{кольквижсяньнымджык}.
The only difference is, however, in the order of markers for case Egr and possessive suffix PxPl1. 
The model is actually giving a correct output, but the input doesn't have all information about the suffix order that the model would need.

There are also instances of word generation where the correct prediction would demand actual lexicographical knowledge, which the model cannot have. 
For example, Komi displays with some nouns an additional stem consonant. It is not possible to predict from the surface form whether this consonant exists and what it is.
So when the model is given input \foreignlanguage{russian}{мек N Sg Ins}, it doesn't predict the correct \foreignlanguage{russian}{мекйӧн} `with a pelt',
but offers the regular but incorrect form 
\foreignlanguage{russian}{мекӧн}.
This is a good example from construction where rule-formulated linguistic knowledge may be necessary for optimal analysis.
It also shows that the model is capable to learn very well the regular structures of the language and does predict them with high accuracy.


\section{Conclusions}

In this paper, we have presented a method for automatically extracting inflectional forms from FST transducers by composing a regular expression transducer for each word with an existing FST transducer. 
This way, we have been able to gather very large morphological training data for analysis, lemmatization and generation for 22 languages, 17 out of which are endangered and fighting for their survival. 
We have used this dataset to train neural models for each language. 
Because the data follows the tags and conventions used in the GiellaLT infrastructure, these neural models can be used directly side by side with the FST transducers in many of the applications that depend on them. 

The results look very good for some languages while being a bit more modest for others. 
Analysis seems to be the hardest problem out of the three, and its training also took the longest time. 
Despite this, many models reached to an over 80\% accuracy in the tasks. 
This is rather good given that the evaluation was conducted entirely on out-of-vocabulary words.

The accuracies reported in this paper are a somewhat lower than what they could be.  This is due to the fact that we ran the evaluation by producing one result only for each input with the neural models and compared that input directly to the one in the test data. As we saw in our analysis, many of the inputs in the test data were ambiguous, which caused the neural model to produce an output that is correct, but not the one in the test data. However, the right way to overcome this problem would be to research how to deal with ambiguity. The neural models we trained can already now produce N best candidates for each input. 

It is probable that within those N best candidates, the models actually cater for the ambiguity and produce other results that are correct as well. For instance, the Finnish word \textit{noita}, can be an accusative singular noun meaning `witch' or a partitive of \textit{nuo} meaning `them'. Knowing how to maximize the number of forms the neural model produces while minimizing the number of incorrect forms is a question for another paper. Although, some methods could already be used with the models trained in this paper by introducing simple modifications to how the results are predicted \cite{silfverberg2019data}.

Even though we aimed for a real world scale morphological tag complexity by querying all possibilities from the FSTs, there are still a couple of morphological categories we did not tackle for practical reasons. One of them is the use of clitics. The problem with these is that they can be attached to almost any kind of word regardless of its part-of-speech and inflectional form. On top of this, multiple clitics can be added one after another. To give an idea of the scale, with clitics, Finnish has 9425 unique forms for nouns (instead of 850), 216 for adverbs (instead of 16), 14794 for adjectives (instead of 1244) and a whopping 88044 forms for verbs (instead of 6667). This means that clitics need to be solved by taking a different approach than the one we had. One could, for example, introduce some forms with different combinations of clitics here and there in the training data, in which case the question arises on how many forms need to appear with clitics in order for the model to generalize their usage.

Compounds and derivations could not be included because of how the FSTs were implemented. 
If you ask an FST for compounds and derivations, you will surely get them! 
Even in such quantities that your computer will run out of RAM and swap memory for the forms of a single word, as there is no limit to how many words can be written together to form a compound or how many times one can derive a new word from another. 
We people might have our cognitive limits for that, but the FSTs will not\footnote{Take, for instance, a look at this derivational Skolt Sami word produced by the FST Piân'njatõõvvõlltâs- ttiatemesvuõt'tsážvuõðtõvvstõlškuät'tteškuättõõlstõlstââst- stõõst\v{c}âtttömâs for piânnai+N+Der/Dimin+Der/N2A+Der/toovvyd +Der/oollyd+Der/jed+V+Der/Caus+Der/Dimin+Der/NomAg+N +Der/Dimin+Der/N2A+Der/teqm+A+Attr+Der/vuott+N +Der/sazh+A+Err/Orth+Attr+Der/vuott+N+Der/toovvyd +Err/Orth+Der/stoollyd+V+Der/shkueqtted+Der/jed +V+Der/Caus+Der/shkueqtted+Der/oollyd+Der/stoollyd +Der/Dimin+V+Der/Dimin+Der/Dimin+V+Der/Dimin+Der/ched +Der/Caus+Der/t+A+Superl+Attr}. 
The problem of compounds is probably best to leave for a separate model to solve, as there are already methods out there for predicting word boundaries \cite{shao2018universal,seeha-etal-2020-thailmcut}. 
The compound splits by such methods could then be fed into the neural models trained in this paper. 
As for derivations, some of them could be included in the training data, but the question of how many forms are needed would still require further research.

\bibliographystyle{acl_natbib}
\bibliography{coling2020}

\end{document}